\documentclass{article}

\usepackage[final]{nips_2017}

\usepackage[utf8]{inputenc} 
\usepackage[T1]{fontenc}    
\usepackage{hyperref}       
\usepackage{url}            
\usepackage{booktabs}       
\usepackage{amsfonts}       
\usepackage{nicefrac}       
\usepackage{microtype}      
\usepackage{times}
\usepackage{latexsym}
\usepackage{color}
\usepackage{tabularx}
\usepackage{wrapfig}

\title{Fast Linear Model for Knowledge Graph Embeddings}

\author{
  Armand Joulin \\
  Facebook AI Research\\
  \texttt{ajoulin@fb.com} \\
   \And
   Edouard Grave \\
  Facebook AI Research\\
  \texttt{egrave@fb.com} \\
   \AND
   Piotr Bojanowski \\
  Facebook AI Research\\
  \texttt{bojanowski@fb.com} \\
   \And
   Maximilian Nickel \\
  Facebook AI Research\\
  \texttt{maxn@fb.com} \\
   \And
   Tomas Mikolov \\
  Facebook AI Research\\
  \texttt{tmikolov@fb.com} \\
}

\begin{document}

\maketitle

\begin{abstract}
  This paper shows that a simple baseline based on a Bag-of-Words (BoW) representation
  learns surprisingly good knowledge graph embeddings.
  By casting knowledge base completion and question answering as supervised
  classification problems, we observe that modeling co-occurences of
  entities and relations leads to state-of-the-art performance with a training time of a few minutes
  using the open sourced library \texttt{fastText}\footnote{Code available at \url{https://github.com/facebookresearch/fastText}}.
\end{abstract}

\section{Introduction}
Learning representations for Knowledge bases (KBs) such as Freebase~\citep{bollacker2008freebase}
has become a core problem in machine learning, with a large variety of
applications from question answering~\citep{yao2014information} to image
classification~\citep{deng2014large}.
Many approaches have been proposed to learn these representations, or embeddings,
with either single relational~\citep{hoff2002latent,perozzi2014deepwalk} or multi-relational
data~\citep{nickel2011three,bordes2013translating}.

These approaches learn graph embeddings by
modeling the relation between the different entities in the graphs~\citep{perozzi2014deepwalk,nickel2016holographic},
Instead, we frame this problem
in a multiclass multilabel classification
problem and model only the co-occurences of entities and relations with a linear classifier based
on a Bag-of-Words (BoW) representation and standard cost functions.
In practice, this approach works surprisingly well on a variety
of standard datasets, obtaining performance competitive with the state-of-the-art approaches
while using a standard text library (i.e., \texttt{fastText}) and running in a few minutes~\citep{joulin2017bag}.

We focus our study on two standard approaches to learn representations for KBs:
knowledge base completion and question answering. For KB completion, our
conclusions extend those of~\citet{kadlec2017knowledge}, that simple models like TransE~\citep{bordes2013translating} work
as well, if not better than more sophisticated ones, if tuned properly.
\citet{kadlec2017knowledge} focus on
a bilinear model designed for KB completion, DistMul~\citep{yang2014embedding}, that still takes a few hours to train on a high-end GPU.
We show that similar performance can be achieved with a linear classifier and a training time reduced to a few minutes.
For question answering, we consider datasets where we have guarantees that the question answer
pairs are covered by the graph in one hop to indirectly learn graph embeddings~\citep{bordes2015large,miller2016key}.
Following~\citet{bordes2014question}, we predict the relation
between the entities appearing in the question and answer pairs to learn embeddings of the graph edges. The embeddings
of the entities, or nodes, are indirectly learned by embedding the questions.
In this setting, we achieve competitive performance as long as we have access
to a clean KB related to the question answering task.

\section{Approach}

\subsection{\texttt{fastText} model}
\label{sec:fasttext}

Linear models~\citep{joachims1998text} are powerful and efficient baselines for text classification.
In particular, the \texttt{fastText} model proposed by~\citet{joulin2017bag}
achieves state-of-the-art performance on many datasets by combining several standard tricks, such as
low rank constraints~\citep{schutze1992dimensions} and n-gram features~\citep{wang2012baselines}.
The same approach can be applied to any problem where the input is a set of discrete
tokens. For example, a KB is composed of entities (or nodes) and relations (or edges) that
can be represented by a unique discrete token.

The model is composed of a matrix~$V$ which is used as a look-up table over the discrete
tokens and a matrix~$W$ for the classifier.  The representations of the
discrete tokens are averaged into BoW representation, which is in turn fed to
the linear classifier.
Using a function $f$ to compute the probability distribution over the classes,
and~$N$ input sets for discrete token (e.g., sentences), leads to minimize:
\begin{eqnarray*}
-\frac{1}{N} \sum_{n=1}^N y_n \log( f (WVx_n)),
\end{eqnarray*}
where~$x_n$ is the normalized BoW of the~$n$-th input set,~$y_n$ the label.
While BoW models are memory inefficient, their memory footprint can be
significantly reduced~\citep{joulin2016fasttext}. The model is trained
asynchronously on multiple CPUs with SGD and a linearly decaying learning rate.

\begin{table*}[t!]
  \centering
\begin{tabular}{lcccc}
\toprule
Method & \multicolumn{2}{c}{WN18} & \multicolumn{2}{c}{FB15k}\\
\midrule
& raw & filtered & raw & filtered \\
\midrule
TransE~\citep{bordes2013translating} & 75.4 & 89.2 & 34.9 & 47.1\\
Rescal~\citep{nickel2012factorizing} & - & 92.8 & - & 58.7 \\
Fast-TransR~\citep{lin2015learning} & 81.0 &  94.6 & 48.8 & 69.8\\
HolE~\citep{nickel2016holographic}  & - & 94.9 & - & 73.9\\
TransE++~\citep{nickel2016holographic}  & - & 94.3 & - & 74.9\\
Fast-TransD~\citep{lin2015learning} & 78.5 & 91.9 & 49.9 & 75.2 \\
\midrule
ReverseModel~\citep{dettmers2017convolutional} & - & 96.9 & - & 78.6\\
HolE+Neg-LL~\citep{trouillon2017complex} & - & 94.7 & - & 82.5 \\
Complex~\citep{trouillon2017knowledge} & - & 94.7 & - & 84.0\\
R-GCN~\citep{schlichtkrull2017modeling} & - & 96.4 & - & 84.2 \\
ConvE~\citep{dettmers2017convolutional} & - & 95.5 & - & 87.3 \\
DistMul~\citep{kadlec2017knowledge} & - & 94.6 & - & 89.3 \\
Ensemble DistMul~\citep{kadlec2017knowledge} & - & 95.0 & - & 90.4\\
IRN~\citep{shen2016implicit} & - & 95.3 & - & 92.7 \\
\midrule
\texttt{fastText} - train & 80.6 & 94.9 & 52.3 & 86.5 \\
\texttt{fastText} - train+valid& 83.2 & 97.6 & 53.4 & 89.9 \\
\bottomrule
\end{tabular}
\caption{Raw and filtered Hit@10 on WN18 and FB15k.
All the numbers are taken from their paper.
Above, methods that should achieve better performance with a
finer hyper-parameter grid, below, methods that were properly tuned.
Higher the better.}\label{tab:kbc}
\end{table*}

\subsection{Loss functions}

We consider two loss functions in our experiments: the softmax function
and a one-versus-all loss function with negative sampling.

\paragraph{Softmax.}
Given $K$ classes, and a score $s_k$ for each class $k$,
the softmax function is defined as $f(s)_k = \exp(s_k)/\sum_{i=1}^K \exp(s_i).$
This function requires the score of every class,
leading to a complexity of $O(Kh)$ where $h$ is the size of the embeddings.
This function is often used to compute the probability distribution of
a finite set of discrete classes.

\paragraph{one-versus-all loss.}
Computing the softmax function over a large number of classes is computationally prohibitive.
We replace it by an independent binary classifier per class, i.e., a set of \emph{one-versus-all} losses.
During training, for each positive example, we draw randomly $k$ negative classes,
and update the $k+1$ classifiers.
The number $k$ is significantly smaller than $K$, reducing the complexity from $O(Kh)$ to $O(kh)$.
This loss has been used for word embeddings~\citep{mikolov2013efficient, bojanowski2017enriching}
as well as to object classification~\citep{joulin2016learning}.

\subsection{Knowledge base completion}

A knowledge base is represented as a set of subject-relation-object triplets
$(e,r,p)$.  Typically, the entity $p$ is predicted
according to the subject $e$ and the relation $r$. With the notations of
the \texttt{fastText} model described in Sec.~\ref{sec:fasttext}, each
entity $e$ is associated with a vector $v_e$ and each relation $r$ with
a vector $v_r$ of the same dimension $h$.
The target entity $p$ is also represented by a $h$ dimensional vector $w_p$.
The scoring function $s_p$ for a triplet $(e,r,p)$ is simply the dot product
between the BoW representation of the input pair $(e,r)$ and the target:
\begin{eqnarray}\label{eq:score}
  s_p(e,r,p) &=& \frac{1}{2} \langle v_e+v_r, w_p \rangle.
\end{eqnarray}
This scoring function does not define a relational model, it only captures
co-occurence between entities and relations.  Additionally, it makes no
assumption about the direction of the relation, i.e., the same relation
embedding is used to predict both ends of a triplet. To circumvent this
problem, we encode the direction in the relation embedding by associating a
relation $r$ with two embeddings, one to predict the subject and one to predict
the object.  While our approach shares many similarities with
TransE~\citep{bordes2013translating}, it differs in several aspects: they use a
ranking loss, their scoring function is an $\ell_2$ distance, and they have one
embedding per entity.  Similarly, if the goal is to predict the relation
between a pair of entities, our scoring function $s_r$ is:
$s_r(e,r,p) = \frac{1}{2} \langle v_e+v_p, w_r \rangle.$
As for entity prediction,  we circumvent the symmetry between subject and
relation by associating each entity with two embeddings, one if the entity is
the subject or the object of a triplet.

\subsection{Question answering}

Question answering problems can be used to learn graph embeddings if framed as
edge prediction problems between entities appearing in the question answer
pairs~\citep{bordes2014question}. The question is represented as a bag of words and
the potential relations are labels. An entity is indirectly represented by the associated
words in the question.

\paragraph{String matching for entity linking.}
The questions and answers are matched to entities in the KB with a string
matching algorithm~\citep{bordes2014question}, using a look-up table between
entities and their string representations.  Every pair of question and answer
in the training set is thus matched to a set of potential pairs of entities.
Several entities are often matched to a question and we use an ad-hoc euristic to
sort them, i.e., using the inverse of their frequency in the training set,
and the size of their associated strings in case of ties (to approximate the
frequency).

\paragraph{Relation prediction for question answering.}
Once a question-answer pair is associated with a set of pairs of entities,
candidate relations are extracted.  Following~\citet{bordes2014question}, we
consider the relations as labels and use \texttt{fastText} to predict them. At
test time, the answer to a question is inferred by taking the most likely
relation and verify if any of the entities matched to the question forms a
valid pair in the KB.  If not, we move to the next most likely relation and
reiterate the process.

\section{Results}

\begin{table*}[t]
  \begin{minipage}[b]{.45\linewidth}
    \begin{center}
  \begin{tabular}{lcc}
\toprule
Method & FB15k-237\\
\midrule
R-GCN~\citep{schlichtkrull2017modeling} &  41.7 \\
DistMul~\citep{yang2014embedding} &  41.9\\
Complex~\citep{trouillon2017knowledge} &  41.9 \\
ConvE~\citep{dettmers2017convolutional} & 45.8 \\
\midrule
\texttt{fastText} - train & 44.8 \\
\texttt{fastText} - train+valid& 45.8 \\
\bottomrule
\end{tabular}
\caption{Filtered Hit@10 on FB15k-237.
The numbers are from~\citet{dettmers2017convolutional}.}\label{tab:237}
\end{center}
\end{minipage}
\hfill
  \begin{minipage}[b]{.45\linewidth}
  \centering
  \begin{tabular}{lc}
  \toprule
  Method  & SVO\\
  \midrule
  TransE~\citep{garcia2015combining} & 70.6 \\
  SME~\citep{bordes2014semantic} & 77.0\\
  LFM~\citep{jenatton2012latent} & 78.0\\
  TATEC~\citep{garcia2015combining} & 80.1 \\
  \midrule
  \texttt{fastText} - train & 79.8 \\
  \texttt{fastText} - train+valid & 79.9 \\
  \bottomrule
\end{tabular}
\caption{Hit$5\%$ on SVO.
The numbers are from~\citet{garcia2015combining}.}\label{tab:svo}
\end{minipage}
\end{table*}
\subsection{Knowledge base completion.}

\paragraph{Datasets.}~We use several standard benchmarks for KB completion:
\begin{itemize}
  \item
    The WN18 dataset is a subset of WordNet, containing 40,943 entities, 18 relation types, and 151,442 triples.
    WordNet is a KB built by grouping synonym words and provides lexical relationships between them.
  \item
    The FB15k dataset is a subset of Freebase, containing 14,951 entities, 1345 relation types, and 592,213 triples.
    Freebase is a large KB containing general facts about the world.
  \item
    The FB15k-237 dataset that is a subset of FB15k with no reversible relations~\citep{toutanova2015representing}.
    It contains 237 relations and 14,541 entities, for a total of 298,970 triples.
  \item
    The SVO dataset is a subset of subject-relation-object triplets extracted from Wikipedia articles, containing $30,605$ entities, $4,547$ relation types and $1.3$M triples.
\end{itemize}

\paragraph{Experimental protocol.}
For WN18, FB15k and FB15k-237, the goal is to predict one end of a triple given the other
end and the relation, e.g., the subject given the object and the
relation. We report Hit@10, also known as Recall@10, on raw and filtered
datasets. Raw means the standard recall measure while filtered means that every
relation that already exists in the KB are first removed, even those in the
test set. The filtered measure allows a direct comparison
of the target entity with negative ones.
On SVO, the goal is to predict the relation given a pair of entities. The measure
is Hit@5$\%$, i.e., Hit@227 for $4,547$ relation types.

\paragraph{Implementation details.}
For both WN18, FB15k and FB15k-237, we use a negative sampling approximation of the softmax
and select the hyper-parameters based on the filtered hits@10 on the validation set.
On WN18 and FB15k,he grid of parameters used
is $[10,25,50,100,150,200]$ for the embedding size $h$, $[100,150,200]$ for the
number of epochs and $[100,200,500]$ for the number of negative examples.
Since FB15k-237 is much smaller, we limit the number of epochs to $[1,5,10]$.
The initial learning rate is fixed at $0.2$.
On WN18, the best set of hyper-parameters are $100$ dimensions, $100$ epochs and $500$ negative samples.
On FB15k, the selected hyper-parameters are $100$ dimensions, $100$ epochs and $100$ negative samples.
On FB15k-237, the best set of hyper-parameters are a hidden of $50$, $10$ epochs and
a $500$ negative samples.
For SVO, the number of relations to predict is quite small, we thus use a full softmax and
select hyper-parameters based on hit@5\%.
The grid of hyper-parameters is $[10,25,50,100,150,200]$ for the embedding size $h$ and
$[1,2,3,4,5]$ for the number of epochs. The initial learning rate is fixed at $0.2$.
For all these experiments, we report both the performance on the model train
on the train set and on the concatenation of the train and validation set, run with the same
hyper-parameters.

\paragraph{Comparison.}
We compare our approach to several standard models in Table~\ref{tab:kbc} on WN18 and FB15k.
We report numbers from their original papers.
Some of them are not using a fine grid of hyper-parameters, which partially explains the gap in performance.
We separate these models from more recent ones for fairer comparison.
Despite its simplicity, our approach is competitive with dedicated pipelines both for
raw and filtered measurements. This extends the findings of~\citet{trouillon2017complex}, i.e.,
the choice of loss function can have a significant impact on overall performance.
Table~\ref{tab:237} extends this observation to a harder dataset, FB15k-237, where our BoW model compares favorably with existing
KB completion models.

We also report comparison on relation prediction dataset SVO in Table~\ref{tab:svo}.
Our approach is competitive with approaches using bigram and high order information, like
TATEC~\citep{garcia2015combining}.
Note TATEC can be, theoretically, used for both
relation and entity prediction, while our model only predicts relations.

\begin{table*}[h]
    \centering
    \begin{tabular}{lcccccc}
      \toprule
      Dataset & WN18 & FB15k & SVO & FB15k-237 & SQ & WikiMovies \\
      \midrule
      Time (sec.) & 165 & 188 & 371 & 28 & 42 & 1 \\
      \bottomrule
    \end{tabular}
  \caption{
Training time for \texttt{fastText} using $20$ threads on a Intel Xeon CPU E5-2680 v3 2.50GHz.}\label{tab:time}
\end{table*}
Table~\ref{tab:time} show the running time for a \texttt{fastText} implementation.
It runs in a few minutes, which is comparable with optimized pipelines like Fast-TransD and Fast-TransR~\citep{lin2015learning}.
Note that similar running times should be achievable for other linear models like TransE.

\subsection{Question answering.}

\paragraph{Datasets.}
We consider two standard datasets with a significant amount of question answer pairs.
\begin{itemize}
  \item
    SimpleQuestion consists of 108,442 question-answer pairs generated from Freebase. It comes with a subset of Freebase with 2M triplets.
  \item
    WikiMovies consists of more than 100,000 questions about movies generated also from Freebase. It comes with a subset of the KB associated with the question-answer pairs.
    This dataset also provides with settings where different preprocessed versions of Wikipedia are considered instead of the KB.
    These settings are beyond the scope of this paper.
\end{itemize}

\paragraph{Implementation details.}
For both SimpleQuestion and MovieWiki, the number of relations are relatively small. We thus use a full softmax.
For SimpleQuestion, the grid of hyper-parameters is $[10, 50, 100, 200]$ for the dimension of the embeddings and $[5,10,50,100]$ for the number of epochs.
We use bigrams and an initial learning rate of $1$.
For MovieWiki, we fixed the embedding size to $16$ since there are only $16$ relations and the number of epochs was selected on the validation set in $[1,5,10,50]$.
We use an initial learning rate of $.3$.

\begin{wraptable}{r}{.5\linewidth}
  \vspace{-10pt}
  \centering
  \begin{tabular}{lc}
    \toprule
    Method & SQ \\
    \midrule
    Random guess~\citep{bordes2015large} & 4.9\\
    CFO~\citep{dai2016cfo}  & 62.6 \\
    MemNN~\citep{bordes2015large} & 62.7 \\
    AMPCNN~\citep{yin2016simple} & 68.3\\
    CharQA~\citep{golub2016character} & 70.9 \\
    CFO + AP~\citep{dai2016cfo} & 75.7\\
    AMPCNN + AP~\citep{yin2016simple} & 76.4\\
    \midrule
    \texttt{fastText} - train  & 72.7 \\
    \texttt{fastText} - train+valid  & 73.0 \\
    \bottomrule
  \end{tabular}
  \caption{Accuracy on the SimpleQuestions dataset~\citep{bordes2015large}.}\label{tab:sq}
  \vspace{-20pt}
\end{wraptable}
\paragraph{SimpleQuestion.}
Figure~\ref{tab:sq} compares this approach with the state-of-the-art.  We learn
a relation classifier with \texttt{fastText} in 42sec.  Using a larger KB,
i.e., FB5M, does not degrade the performance, despite having much more
irrelevant entities.  Our approach compares favorably well other with question
answering systems. This suggests that the learned embeddings capture some
important information about the KB.  Note, however, that the performance is
very sensible to the quality of the entity linker and the ad-hoc sorting of
extracted subjects. Typically, going from a random order to the one used in
this paper gives a boost of up to $10\%$ depending on the hyper-parameters.

\paragraph{WikiMovies.}
Table~\ref{tab:movieqa} compares our models with several state-of-the-art
pipelines. In the case where the clean KB is accessible, our method works very
well.
\texttt{fastText} runs in 1sec. for relation prediction.
Note that this dataset was primarily made for the case where only text is
available. This setting goes beyond the scope of our method, while a more
general approach like KV-memNN still works reasonably
well~\citep{miller2016key}.
\begin{table*}[h]
  \centering
  \begin{tabular}{lccccc}
\toprule
Method &
SE &
MemNN &
QA System &
KV-MemNN &
\texttt{fastText} \\
\midrule
WikiMovies & 54.4 & 78.5 & 93.5 & 93.9 & 95.9 \\ 
\bottomrule
\end{tabular}
\caption{Test result (\% hits@1) on the WikiMovies dataset with the full KB.
The numbers are taken form~\citet{miller2016key}.
QA System refers to \citet{bordes2014question}.}\label{tab:movieqa}
\end{table*}

\section{Conclusion}
In this paper, we show that linear models learn good
embeddings from a KB by recasting graph related problems into supervised
classification ones.  The limitations of such approach are that it requires a
clean KB and a task that uses direct information about local connectivity in
the graph.  Moreover, the observation that our non-relational approach provides
state-of-the-art performance on KBC benchmarks raises also important questions
regarding the evaluation of link-prediction models and the design of benchmarks
for this task.

\paragraph{Acknowledgement.}
We thank Timoth\'ee Lacroix, Nicolas Usunier, Antoine Bordes and the rest of
FAIR for their precious help and comments. We also would like to thank Adam
Fisch and Alex Miller for their help regarding MovieWiki.

\bibliography{nips_2017}
\bibliographystyle{emnlp_natbib}
\end{document}